\documentclass[lettersize,journal]{IEEEtran}
\usepackage[caption=false,font=normalsize,labelfont=sf,textfont=sf]{subfig}

\usepackage{float}
\usepackage{xspace}
\usepackage{graphicx} \graphicspath{{figures/}} 
\usepackage{amsmath,amssymb,nicefrac,bbm,pifont}
\usepackage{cite}
\usepackage[linesnumbered,ruled,vlined]{algorithm2e}
\usepackage{acronym}
\usepackage{subfig}
\usepackage{booktabs,tabularx,colortbl,multirow,multicol,array,makecell,tabularray}
\usepackage{overpic}
\usepackage[misc]{ifsym}
\usepackage{enumitem}
\usepackage{orcidlink}
\usepackage{dblfloatfix}
\usepackage{siunitx}
\usepackage{xr}
\usepackage{verbatim}
\usepackage{textcomp}
\usepackage{dsfont}
\usepackage{diagbox}
\usepackage{setspace}
\usepackage{hyperref}
\usepackage{cleveref}
\usepackage{wrapfig}
\usepackage[utf8]{inputenc} 
\usepackage[T1]{fontenc}    
\usepackage{url}            
\usepackage{amsfonts}       
\usepackage{microtype}      
\usepackage{xcolor}         
\usepackage{array}
\newcolumntype{C}[1]{>{\centering\arraybackslash}p{#1}}
\usepackage{tcolorbox}
\usepackage{etoc}
\tcbuselibrary{breakable,skins}
\newcommand{\cmark}{\ding{51}}
\makeatletter
\DeclareRobustCommand\onedot{\futurelet\@let@token\@onedot}
\def\@onedot{\ifx\@let@token.\else.\null\fi\xspace}
\def\eg{\emph{e.g}\onedot}

\makeatother

\definecolor{pastelblue}{HTML}{2e54a1}
\definecolor{pastelgreen}{HTML}{588e31}
\definecolor{pastelorange}{HTML}{c65f11}
\definecolor{pastelpink}{HTML}{c00000}
\definecolor{GOA}{HTML}{155636}
\definecolor{OORA}{HTML}{1d275b}
\definecolor{resultGreen}{HTML}{588e31} 
\definecolor{resultRed}{HTML}{c00000}

\acrodef{ai}[AI]{Artificial Intelligence}
\acrodef{sgg}[SGG]{Scene Graph Generation}
\acrodef{sga}[SGA]{Scene Graph Anticipation}
\acrodef{lsa}[LSGA]{Linguistic Scene Graph Anticipation}
\acrodef{llm}[LLM]{Large Language Model}
\acrodef{gags}[GAGS]{Grounded Action Genome Scenes}
\acrodef{ootsm}[OOTSM]{Object‑Oriented Two‑Stage Method}
\acrodef{lora}[LoRA]{Low-Rank Adaptation}
\acrodef{goa}[GOA]{Global Object Anticipation}
\acrodef{oora}[OORA]{Object-Oriented Relationship Anticipation}

\begin{document}

\title{Language-Driven Object-Oriented Two-Stage Method for Scene Graph Anticipation}

\author{%
Xiaomeng~Zhu\,\orcidlink{0000-0002-0704-6314},
Changwei~Wang\,\orcidlink{0000-0001-8259-7717},
Haozhe~Wang\,\orcidlink{0000-0002-9299-6305},
Xinyu~Liu\,\orcidlink{0009-0007-0456-921X},
Fangzhen~Lin\,\orcidlink{
0000-0002-3141-8675}
\thanks{
Manuscript received December x, 2025; revised x x, 2025.
}
\thanks{
Xiaomeng Zhu, Haozhe Wang, and Fangzhen Lin are with the Department of Computer Science and Engineering, The Hong Kong University of Science and Technology, Hong Kong 999077, China (email: xiaomeng.zhu@connect.ust.hk, hwangha@connect.ust.hk, flin@cse.ust.hk).}
\thanks{
Changwei Wang is with the Key Laboratory of Computing Power Network and Information Security, Ministry of Education, Shandong Computer Science Center, Qilu University of Technology, China (email: changweiwang@sdas.org).}
\thanks{
Xinyu Liu is with the Academy of Interdisciplinary Studies, The Hong Kong University of Science and Technology, Hong Kong 999077, China (email: xliugd@connect.ust.hk).}
\thanks{
Implementable code is available at: \url{https://github.com/ZhuXMMM/OOTSM}.}
}

\maketitle

\begin{abstract}
A scene graph is a structured representation of objects and their spatio-temporal relationships in dynamic scenes. \ac{sga} involves predicting future scene graphs from video clips, enabling applications in intelligent surveillance and human--machine collaboration.
While recent \ac{sga} approaches excel at leveraging visual evidence, long-horizon forecasting fundamentally depends on semantic priors and commonsense temporal regularities that are challenging to extract purely from visual features.
To explicitly model these semantic dynamics, we propose \textbf{\ac{lsa}}, a linguistic formulation of \ac{sga} that performs temporal relational reasoning over sequences of textualized scene graphs, with visual scene-graph detection handled by a modular front-end when operating on video. Building on this formulation, we introduce \textbf{\ac{ootsm}}, a language-based framework that anticipates object-set dynamics and forecasts object-centric relation trajectories with temporal-consistency regularization, and we evaluate it on a dedicated benchmark constructed from Action Genome annotations.
Extensive experiments show that compact fine-tuned language models with up to 3B parameters consistently outperform strong zero- and one-shot API baselines, including GPT-4o, GPT-4o-mini, and DeepSeek-V3, under matched textual inputs and context windows. When coupled with off-the-shelf visual scene-graph generators, the resulting multimodal system achieves substantial improvements on video-based \ac{sga}, boosting long-horizon mR@50 by up to 21.9\% over strong visual \ac{sga} baselines.
\end{abstract}


\section{Introduction} \label{sec:introduction}
Understanding relationships in human--scene interactions requires more than recognizing static information in a single frame; it necessitates comprehending complex interaction patterns that evolve among objects over time~\cite{mohamed2020social,wang2021spatio}. Recently, spatio-temporal scene graphs have emerged as a powerful structured representation of these dynamics by decomposing a scene into pairwise relations between humans and objects~\cite{xu2017scene,ji2020action,li2022dynamic,qiu2023virtualhome}. Because they already encode how interactions change over time, extending these graphs into the future is a valuable next step~\cite{liu2020forecasting,peddi2024towards}. This predictive capability underpins diverse applications—for instance, anticipating abnormal behaviours in intelligent-surveillance footage \cite{sarker2021semi,chiranjeevi2024anomaly} or inferring user intent early in human–machine collaboration \cite{zhang2024multi,ozdel2024gaze}.

To study this predictive problem more concretely, we therefore turn to focus on the recently proposed \ac{sga} challenge. Given a sequence of observed video frames, \ac{sga} aims to predict future scene graphs that capture both spatial and temporal relationships among objects. Despite its promise, existing \ac{sga} methods draw almost exclusively on visual cues contained in spatio-temporal scene graphs, limiting their ability to incorporate rich semantic priors for interaction prediction~\cite{yu2023visually,li2025unbiased,hsieh2025generation}, which can be vital for nuanced and long-horizon forecasting.
This observation leads to a central question: how can we effectively capture and predict human-object interaction dynamics by combining visual evidence with commonsense priors?

Current approaches to \ac{sga} mostly utilize transformer-based architectures that jointly model temporal dependencies and spatial relationships from visual inputs~\cite{cong2021spatial,li2022dynamic,mazzia2022action,feng2023exploiting}.
This joint optimization often entangles pixel level recognition with long-horizon relational reasoning, making it difficult to inject (or even measure) the role of semantic priors and commonsense regularities~\cite{chen2014enriching,vedantam2015learning,wang2020visual,heo2023simple,wu2023grounded,wang2025vl,wang2025emergent,zhao2025m3w}. 
Therefore, we argue that an effective strategy for \ac{sga} is to explicitly decouple visual perception from semantic reasoning, organizing them within a unified multimodal framework: a video scene graph generator serves as the perception front-end, while a language-based module performs temporal relational reasoning on top of the extracted graphs. This separation allows advances in either vision or language components to be integrated without retraining the entire pipeline, and aligns with the modular design commonly adopted in multimedia understanding systems~\cite{chen2014enriching,vinyals2015show,antol2015vqa,gao2017video}.

Based on this decoupled multimodal strategy, we develop a modular \ac{sga} framework where existing scene-graph detectors extract visual representations while a language model reasons over temporal graph dynamics. We formalize \ac{lsa} to enable interpretable reasoning within this architecture. Specifically, a \ac{lsa} task involves predicting future scene graphs given a sequence of scene graphs corresponding to a video clip. By conducting its core anticipation reasoning primarily in the language domain, \ac{lsa} uniquely positions \acp{llm} to apply their extensive linguistic knowledge and understanding of the learned world~\cite{yuan2023gpt,zhang2023video,shi2024commonsense,qian2024streaming,wen2025enhanced} to the \ac{sga} challenge. Our main contributions in this work are as follows:
\begin{enumerate}[leftmargin=1.2em, label=\textbullet]
\item We formalize \ac{lsa} as a linguistic formulation of \ac{sga} and construct a dedicated benchmark by converting Action Genome into temporally ordered scene-graph sequences with observed/predicted splits under the standard \ac{sga} protocol.
\item We propose \ac{ootsm}, an object-oriented two-stage framework that anticipates future object sets and refines object-centric relational trajectories through temporal-consistency regularization.
\item We develop a unified prompting and fine-tuning strategy that adapts compact open-source \acp{llm} to \ac{lsa} without requiring visual features. A 3B-parameter Llama model trained with this strategy outperforms strong API baselines including GPT-4o by 9.6\% in R@50 and 3.2\% in mR@50 under matched textual inputs and context windows.
\item When integrated with frozen scene-graph detectors, \ac{ootsm} improves video-based \ac{sga} performance, achieving up to 21.9\% mR@50 gain on long-horizon prediction while maintaining robustness under detection noise.
\end{enumerate}

The remainder of this paper is organized as follows. \Cref{sec:related_work} reviews existing scene graph methods. \Cref{sec:method} presents our two-stage framework for linguistic scene graph anticipation with object dynamics modeling. \Cref{sec:experiment} provides comprehensive evaluation on text-based and video-based settings. \Cref{sec:conclusion} concludes the paper.

\section{Related Work}
\label{sec:related_work}
\subsection{Scene Graph Generation.}  
\label{sec:related_work_sgg}
\ac{sgg} characterizes images through object-predicate graphs. Early approaches utilized frequency biases \cite{zellers2018neural} or graph neural networks \cite{xu2017scene} on Visual Genome \cite{krishna2017visual}, with later work addressing long-tail relations via contextual attention and debiasing strategies \cite{tang2020unbiased}. 

Recent advances in multimedia have tackled inherent challenges in scene graph understanding through multiple complementary approaches. Zero-shot predicate prediction methods leverage knowledge priors to handle unseen relations \cite{li2022zero}. Scene graph refinement networks enhance graph quality through structured reasoning for downstream tasks like visual question answering \cite{qian2022scene}. For temporal scenarios, end-to-end video scene graph generation frameworks unify object detection, tracking, and relation recognition through transformer-based architectures \cite{zhang2023end}. Attention-guided relation detection approaches further improve performance by modeling distinctions among different visual cues in dynamic scenes \cite{cao2021attention}. Additionally, debiasing methods that address visual and semantic imbalances have been shown to mitigate class skew in both static and video settings \cite{li2025unbiased}. Our framework accepts \ac{sgg} outputs as input, ensuring compatibility with any state-of-the-art detector and making improvements orthogonal to our method.

\subsection{Scene Graph Anticipation.}
\label{sec:related_work_sga}

\ac{sga} extends \ac{sgg} to temporal prediction over video streams. Current methods employ diverse architectural strategies: spatio-temporal transformers capture both spatial context and temporal dependencies \cite{cong2021spatial,li2022dynamic}, spatial-temporal attention mechanisms enhance feature representations across frames \cite{yan2019stat}, and continuous latent trajectory models via ODEs/SDEs characterize object dynamics in latent space \cite{peddi2024towards,peddi2025towards}. Probabilistic frameworks generate multiple plausible futures to account for prediction ambiguity \cite{hu2020probabilistic}, while object-centric approaches decompose scenes into individual entities for prediction \cite{chen2019uni}. Additionally, spatio-temporal memory networks consolidate information across long temporal ranges for improved consistency \cite{lu2023vmemnet}. Most approaches assume object persistence and rely primarily on visual cues for dynamics modeling \cite{peddi2024towards}. Our work complements these efforts by leveraging language models' commonsense reasoning capabilities, enabling explicit modeling of object set dynamics through semantic understanding without requiring dense visual information.

\subsection{Large Language Models for Scene-Graph Tasks.}

Recent work leverages \acp{llm} and vision-language models to enhance scene graph understanding and generation. For static scene graphs, SDSGG \cite{chen2024scene} employs \acp{llm} to generate scene-specific descriptions with role-playing mechanisms for open-vocabulary prediction, improving adaptation to visual context. VLPrompt \cite{zhou2023vlprompt} leverages natural-language prompts to improve predictions for rare relations by encoding linguistic priors. LLM4SGG \cite{kim2024llm4sgg} parses natural language descriptions into structured scene graph triples, enabling weakly supervised learning from textual annotations. Commonsense validation approaches utilize small-scale \acp{llm} to filter out commonsense-violated predictions, complementing traditional detection models \cite{jiang2025enhancing}. For temporal understanding, vision-language models such as Video-LLaMA \cite{zhang2023video} and HyperGLM \cite{nguyen2024hyperglm} demonstrate multimodal reasoning capabilities, though these primarily focus on description and question-answering rather than future prediction. Scene graph parsing from natural language descriptions has been explored through \ac{llm}-based methods that convert textual inputs into structured scene graphs \cite{yang2025llm}. Our approach uniquely employs \acp{llm} as the exclusive reasoning engine for long-term \ac{sga}.

\begin{figure*}[t!]
    \centering
    \includegraphics[width=\linewidth]{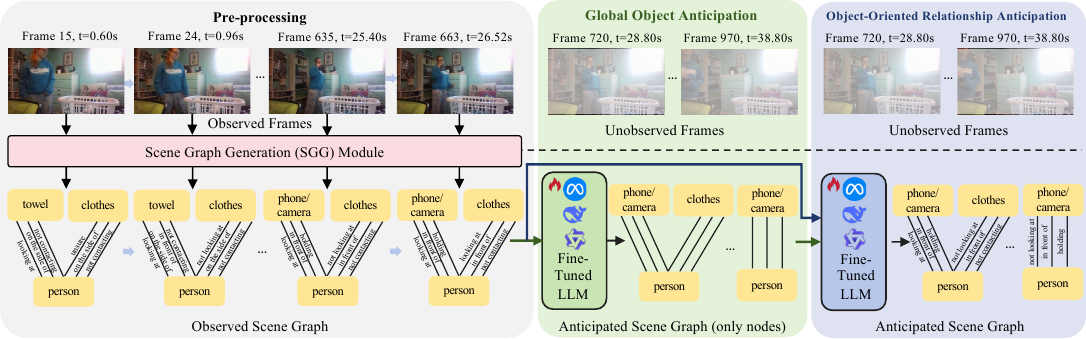}
    \caption{\textbf{Overall \ac{ootsm} pipeline.} 
    The left branch converts observed video frames into compact textual scene graphs; \textcolor{GOA}{\textbf{GOA}} module performs dynamic object anticipation, whereas \textcolor{OORA}{\textbf{OORA}} module carries out object-oriented relationship anticipation. 
    Additionally, scene graphs can bypass \textcolor{GOA}{\textbf{GOA}} for direct input into \textcolor{OORA}{\textbf{OORA}} (dotted arrow), provided the continuous-object constraint is maintained—only objects present in the final observed frame are projected forward.
    The dashed gray path is optional visual \ac{sgg} tool integration.}
    \label{fig:pipeline}
\end{figure*}

\section{Method}
\label{sec:method}

In this section, we formalize the \ac{lsa} task in \Cref{sec:preliminary}. Then we introduce the motivation and design principles of our two-stage architecture in \Cref{sec:framework}. We detail the \ac{goa} module and \ac{oora} module in \Cref{sec:global} and \Cref{sec:refinement}, respectively. Finally, we describe the multimodal integration pathway for end-to-end video-based \ac{sga} in \Cref{sec:integration}.

\subsection{Preliminary}
\label{sec:preliminary}

Generally, the classical \ac{sga} setting takes a raw video prefix $F_{1:n}$ as input and requires a model to return a set of predicted scene graphs $\hat G_{n+1:T}$ that describe the objects and their pairwise relationships in the remaining frames.  In practice, state-of-the-art pipelines realise this mapping with a single multi-modal network that must simultaneously decode visual patterns and reason over semantic structures—a design that conflates pixel-level recognition with symbolic anticipation.

We propose \ac{lsa} to effectively disentangle these two complementary aspects. Specifically, an external scene-graph detector first transforms the observed frames into a sequence of graphs $G_{1:n}$. The anticipation module then predicts $\hat{G}_{n+1:T}$ solely from this graph sequence, operating with its core anticipation reasoning entirely within the language domain and thereby leveraging the commonsense priors embedded in \acp{llm}. Formally, while \ac{sga} learns a mapping $F_{1:n}\!\mapsto\!\hat{G}_{n+1:T}$, \ac{lsa} concentrates on the semantic sub-problem $G_{1:n}\!\mapsto\!\hat{G}_{n+1:T}$; consequently, the visual front-end can be substituted without necessitating retraining of the reasoning component.

Throughout the paper, we denote the unique set of objects (each coinciding with a single category) by $\mathcal{O}=\{o_1,\dots,o_{N_O}\}$ and the set of relation types by $\mathcal{R}=\{r_1,\dots,r_{N_R}\}$.  Following \cite{ji2020action,peddi2024towards}, $\mathcal{R}$ is partitioned into three disjoint subsets: attention relations $\mathcal{R}_{\text{attn}}$, spatial relations $\mathcal{R}_{\text{spat}}$, and contact relations $\mathcal{R}_{\text{cont}}$.  Each graph is therefore represented as a collection of triples $(\text{human},\,o,\,r)$ with $o\in\mathcal{O},\,r\in\mathcal{R}$, providing a structured textual input that can be processed by any language model while remaining compatible with existing video annotations used later in the paper.

\subsection{Overall Framework}
\label{sec:framework}
Existing \ac{sga} methods often assume continuous object sets between observed and future frames, yet this assumption substantially limits their performance. Analysis on Action Genome reveals that only 61\% of videos maintain identical object sets, with 39\% exhibiting object dynamics (14\% new appearances, 25\% disappearances). An oracle experiment demonstrates that strict adherence to last-frame objects limits Recall@10 to 67\%, leaving a 33\% performance gap. These findings motivate explicit future object anticipation.

To enable explicit object anticipation while addressing context window constraints of small-sized \acp{llm}, we propose \ac{ootsm}, a two-stage framework for text-based \ac{lsa} that leverages \ac{llm} for future scene graph prediction (\cref{fig:pipeline}). This framework first generates a global forecast of object occurrences through \ac{goa}, capturing object set dynamics beyond the last observed frame, followed by targeted object-oriented prompts in \ac{oora} for relation refinement—each stage remaining within token limitations. To optimize context utilization, we employ temporal graph merging, combining consecutive frames with identical relationships as follows:
\begin{equation}
\label{eq:merge}
G(F_t) \equiv G(F_{t+1})
\end{equation}
\noindent where $G(F_t)$ represents the scene graph of frame $t$. This aggregation focuses the model's attention on meaningful scene transitions rather than redundant patterns.

Expanding on this approach, we now detail the two-stage mechanism through which our anticipation framework operates:
(i) First, \textcolor{GOA}{\textbf{\ac{goa}}} module utilizes a fine-tuned LLM to generate holistic forecasts of object occurrences and their general relational patterns, though this global approach may not capture precise temporal transitions between individual relationships.
(ii) Second, \textcolor{OORA}{\textbf{\ac{oora}}} module addresses context window constraints by generating targeted per-object prompts, enabling the model to refine each object's relational trajectory through multi-label classification with temporal continuity regularization, thus yielding more accurate relation predictions across the prediction horizon.

The final integration stage consolidates these object-level predictions into a coherent, temporally consistent scene graph, resolving potential conflicts and ensuring realistic interaction trajectories. For applications involving raw visual data, we provide an optional integration pathway leveraging existing \ac{sgg} methodologies, such as STTran~\cite{cong2021spatial}, to bridge visual detections with our semantic anticipation capabilities, thereby enhancing applicability and interpretability across multimodal domains.

\subsection{Global Object Anticipation} \label{sec:global}
\begin{figure}[t!]
    \centering
    \includegraphics[width=\linewidth]{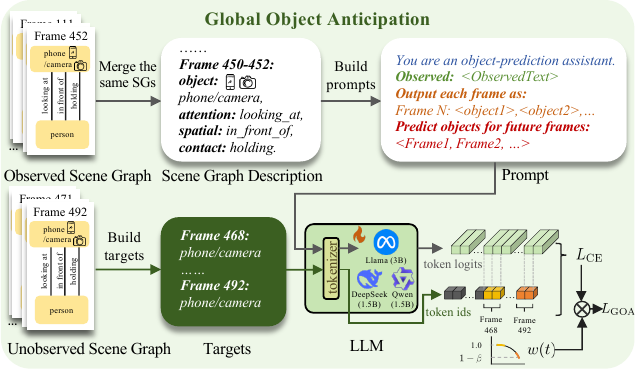}
    \caption{\textbf{GOA training flow.}
    Observed scene graphs with identical structures are first merged and converted to textual descriptions, then combined with instructions to construct prompts. A finetuned LLM subsequently predicts future object sets, supervised by unobserved scene graph targets via temporally weighted cross-entropy loss $L_{\text{GOA}}$.}
    \label{fig:Stage0}
\end{figure}

In the \ac{goa} module, we leverage the generative capabilities of a fine-tuned \ac{llm} to anticipate future scene graphs with a global perspective, as illustrated in \cref{fig:Stage0}. Given observed video segments consolidated into coherent textual blocks ${G}_1, \dots, {G}_n$, we construct a comprehensive prompt structured as:
\begin{equation}
P = H \oplus \text{ObservedText} \oplus \text{Instruction} \oplus \text{FutureFrameInfo}
\end{equation}
\noindent where, $\oplus$ denotes string concatenation. \textcolor{pastelblue}{$H$} provides task-oriented context,
\textcolor{pastelgreen}{\text{ObservedText}} encapsulates the temporal sequence of observed frames,
\textcolor{pastelorange}{\text{Instruction}} delineates precise output formatting requirements,
and \textcolor{pastelpink}{\text{FutureFrameInfo}} explicitly enumerates the future frames requiring predictions.

Conventional uniform weighting treats all future frames equally, yet our experiments (\cref{tab:ablation_study}) show this approach compromises near-term accuracy as uncertain long-horizon predictions introduce high-variance gradients. We implement a cosine attenuation schedule for distant frames, reflecting their diminished signal-to-noise ratio and enabling the optimizer to prioritize higher-fidelity supervision. The loss for \ac{goa} is formulated as:
{\begin{equation}
L_{\text{GOA}}
= \frac{\sum_{t=n+1}^{T} w(t)\,\sum_{i=1}^{K_t}L_{\text{CE}}\bigl(p_{t,i},y_{t,i}\bigr)}
       {\sum_{t=n+1}^{T} w(t)\,K_t}\,,
\label{eq:Stage0_loss}
\end{equation}}
where $t$ ranges over the future graphs $n+1,\dots,T$, $w(t)$ is the temporal weight, and for each graph, $K_t$ denotes the number of byte-pair-encoded tokens in the target string representation of graph $G_t$, with predicted probabilities $p_{t,i}$ and ground‐truth labels $y_{t,i}$ for $i=1,\dots,K_t$. The weighting function $w(t)$ is defined as:
{\begin{equation}
w(t)
= \beta\Bigl[1 + \cos\Bigl(\pi\,\frac{t-(n+1)}{T-(n+1)}\Bigr)\Bigr] + (1-\beta)\,,
\label{eq:cosine_weight}
\end{equation}}
where \(\beta\in[0,1]\) balances the base weight and cosine modulation.
 
\ac{goa}'s predictions identify object occurrences and relational contexts temporally. Despite establishing global coherence, insufficient temporal granularity precludes precise scene graph construction, necessitating \ac{oora} module refinements.

\begin{figure*}[t]
    \centering
    \includegraphics[width=\linewidth]{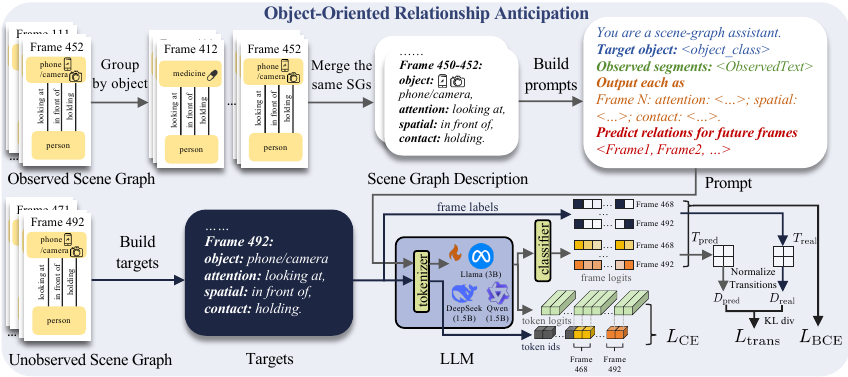}
    \caption{\textbf{OORA training flow.} 
    Observed scene graphs are first grouped by object, then frames with identical scene graph structures are merged. Scene graph descriptions are combined with object-specific information to build prompts. These object-specific prompts guide a finetuned LLM to predict future relationships ($L_{\text{CE}}$). An auxiliary classifier generates per-frame relationship probabilities ($L_{\text{BCE}}$), while a transition regularizer ($L_{\text{trans}}$) ensures temporal coherence by penalizing improbable state transitions.}
    \label{fig:Stage1}
\end{figure*}

\subsection{Object-Oriented Relationship Anticipation}
\label{sec:refinement}
\ac{oora} of our framework addresses a fundamental limitation of \ac{goa},
as illustrated in \cref{fig:Stage1}: while \ac{goa} provides comprehensive scene-level forecasts, it lacks precision in tracking individual objects' temporal evolution. To resolve this, we implement object-oriented refinement, generating detailed multi-frame relationship distributions for each target entity. 
Beginning with the globally predicted scene graph $\widehat{G}^{(0)}$, we isolate the set of predicted future objects $\mathcal{O}$. For each object $o \in \mathcal{O}$, we craft an object-specific prompt:
\begin{equation}
P_o = H_o \oplus \text{ObservedSegments}_o \oplus \text{Instruction}_o \oplus \text{FutureFrames}_o
\end{equation}
\noindent where
\textcolor{pastelblue}{$H_o$} provides the object-centric prediction context,
\textcolor{pastelgreen}{\text{ObservedSegments}$_o$} succinctly summarizes the object’s historical presence and interactions,
\textcolor{pastelorange}{\text{Instruction}$_o$} explicitly specifies the expected format of relationship predictions,
and \textcolor{pastelpink}{\text{FutureFrames}$_o$} enumerates the prediction horizon, enabling targeted temporal predictions.

Relationship predictions in sequential frames often exhibit implausible transitions and temporal inconsistencies, severely compromising long-term forecasting reliability. To address this critical challenge, we introduce a temporal-transition regularizer $L_{\text{trans}}$ that penalizes improbable relationship state transitions across consecutive frames. Concretely, we first transform the frame-wise transition counts $T_{\text{real}}^r$ and the soft predictions $T_{\text{pred}}^r$ into normalized histograms $D_{\text{real}}^r$ and $D_{\text{pred}}^r$. To ensure robustness against sparse relationships and minor fluctuations, we apply gating mechanisms: a count threshold $\delta$ filters relationships with insufficient observed transitions, while a probability-change threshold $\tau=0.2$ excludes trivial variations in consecutive predictions. The discrepancy between distributions is measured using symmetrized Kullback–Leibler divergence:
{\begin{equation}
L_{\mathrm{trans}}
= \frac{1}{N_R}\sum_{r=1}^{N_R}
\frac{1}{2}\left(\mathrm{KL}\left(D_{\mathrm{pred}}^r \,\|\, D_{\mathrm{real}}^r\right)
+\mathrm{KL}\left(D_{\mathrm{real}}^r \,\|\, D_{\mathrm{pred}}^r\right)\right),
\end{equation}}
Here, $T_{\text{real}}^r$ and $T_{\text{pred}}^r$ represent frame-wise relationship transition frequencies for ground-truth and predicted probabilities, respectively. This temporal-consistency prior discourages unnecessary high-frequency changes in relation trajectories and reduces error accumulation at long horizons.

To compute transition histograms, we employ an auxiliary discriminative classification layer trained with Binary Cross-Entropy loss $L_{\text{BCE}}$, which maps internal representations to relationship category probabilities.

Combining these discriminative losses with the generative teacher-forcing cross-entropy loss $L_{\text{CE}}$, we form the complete \ac{oora} training objective:
{\begin{equation}
L_{\text{OORA}} = L_{\text{CE}} + L_{\text{BCE}} + \lambda L_{\text{trans}},
\end{equation}}
where $L_{\mathrm{CE}}$ denotes the teacher-forcing cross-entropy on the next-token distribution,
$L_{\mathrm{BCE}}$ supervises an auxiliary classification head that is trained independently of the frozen \ac{llm} parameters,  
and $L_{\mathrm{trans}}$ enforces temporal smoothness of relationship transitions.  
A single hyper-parameter $\lambda$ controls the trade-off between temporal
regularization and the two accuracy-oriented terms; we set $\lambda = 0.03$ in experiments.

\subsection{Integration Framework for Video \ac{sga}}
\label{sec:integration}
We propose an optional integration pathway with established \ac{sgg} tools to extend our textual \ac{sga} framework to raw video inputs, bridging visual perception and semantic anticipation. The process begins with frame-level detection, where \ac{sgg} tools extract bounding boxes, object labels $\in \mathcal{O}$, and relationships $\in \mathcal{R}$, which are converted into our standardized textual format.

We implement temporal merging to consolidate consecutive frames with identical elements, optimizing for textual compactness without semantic redundancy. This condensed ``observed text'' serves as input to \ac{goa}'s prompt for initial object anticipation, consistent with our primary pipeline. The anticipation then follows our two-stage methodology: the \ac{goa} module generates holistic future forecasts for the first stage while \ac{oora} provides object-centric refinements for the next stage.

Finally, we establish bidirectional mapping between textual predictions and spatial representations by linking semantic relationships to their corresponding bounding box indices. This produces a comprehensive spatiotemporal scene graph with both semantic and spatial localization data, enabling end-to-end video-based \ac{sga}. To validate robustness, we evaluate \ac{sgg} noise impact using ground-truth scene graphs (seen in \cref{tab:pure_text_sga}) and detector outputs (seen in \cref{tab:sga_gags_constraint}). Additionally, to encourage high-confidence relation scores and suppress noise, we augment \ac{sgg} training with a margin-based multilabel-threshold loss. Specifically, for predicted probabilities $\mathbf{p}$ and ground-truth labels $\mathbf{y}$, we define:
\begin{multline}
{L}_{\text{thr}}(\mathbf{p},\mathbf{y})
=\frac{1}{N_R}\sum_{r=1}^{N_R}
\Bigl[
y_r\,\max\bigl(0,\gamma_{\text{pos}}-p_r\bigr) \\
+(1-y_r)\max\bigl(0,p_r-\gamma_{\text{neg}}\bigr)
\Bigr],
\end{multline}
where $\gamma_{\text{pos}}$ and $\gamma_{\text{neg}}$ are margin parameters. The final objective combines ${L}_{\text{rel}} = {L}_{\text{BCE}} + \eta\,{L}_{\text{thr}}$, creating an explicit decision margin that reduces ambiguous outputs. In our experiments, we set $\gamma_{\text{pos}}=0.9$, $\gamma_{\text{neg}}=0.5$, and $\eta=0.5$.

\section{Experiment}
\label{sec:experiment}
We conduct comprehensive evaluations on the standard Action Genome dataset across text-based \ac{lsa} and video-based \ac{sga} settings. Our experiments include baseline comparisons in \Cref{exp:text_lsga_results,exp:sga_results}, ablation studies on key components in \Cref{exp:abl_weight_trans,exp:temporal_reasoning,exp:one_shot_results}, hyperparameter analysis in \Cref{sec:obs_len,exp:weighting}, and robustness evaluation in \Cref{exp:robutness_analysis}.

\subsection{Dataset and Evaluation Protocol}
\label{exp:dataset_protocol}
We evaluate on Action Genome (AG)~\cite{ji2020action}, the standard benchmark for scene graph anticipation with dense frame-level human-object interaction annotations. Following the established \ac{sga} protocol~\cite{peddi2024towards}, we filter videos with fewer than three annotated frames, resulting in 11.4K videos, and adopt the official train-test split. The dataset comprises 35 object classes and 25 relation classes (attention, spatial, contact). AG remains the only real-world video dataset with comprehensive annotations for temporal scene graph reasoning, and all prior \ac{sga} methods evaluate exclusively on this dataset for fair comparison.

Evaluation employs R@K and mR@K metrics with $K\in\{10, 20, 50\}$. R@K measures anticipation of relevant relationships between observed objects, while mR@K addresses class imbalance by equally weighting predicate performance. We vary observation ratios $\mathcal{F} \in \{0.3, 0.5, 0.7, 0.9\}$ to assess short-term and long-term capabilities. We adopt the standard ``With Constraint'' setting used in VidSGG and \ac{sga}, where each subject-object pair is assigned at most one predicted predicate (the highest-scoring one), and R@K/mR@K are computed over the top-$K$ scoring triplets per frame. This aligns with \acp{llm}' discrete categorical output nature, unlike global ranking schemes that assume graded confidence distributions.

\subsection{Experimental Setup}
\label{exp:setup}
For pure textual \ac{lsa}, predictions use ground-truth annotations from observed frames. We compare \ac{ootsm} against two groups: (1) API-based baselines using direct inference without fine-tuning (GPT-4o, GPT-4o-mini, DeepSeek-V3), and (2) fine-tuned LLM variants with alternative backbones (Llama-3.2-3B, DeepSeek-R1-1.5B, Qwen 2.5-1.5B) integrated within our framework. This text-based setting directly validates our linguistic paradigm for temporal reasoning, decoupled from visual perception challenges.

For video-based \ac{sga}, we evaluate in the GAGS setting where ground-truth objects and bounding boxes are provided by Action Genome annotations. This isolates temporal reasoning from object detection challenges. We benchmark against state-of-the-art methods: STTran~\cite{cong2021spatial}, DSGDet~\cite{feng2023exploiting}, SceneSayerODE, and SceneSayerSDE~\cite{peddi2024towards}.

All experiments are conducted on a single NVIDIA A100 GPU with 40GB memory. We employ LoRA-based fine-tuning with rank=32 and $\alpha$=32, training the \ac{goa} module for 5 epochs and \ac{oora} module for 10 epochs using SGD optimization with learning rate $1\times10^{-5}$ and batch size 1. The context window is set to 2048 tokens to balance temporal coverage with computational efficiency. During inference, we apply nucleus sampling with temperature=0.7 and top-p=0.4 to maintain generation diversity. Output length constraints are adapted to observation ratios: for $\mathcal{F}\in\{0.3, 0.5, 0.7, 0.9\}$, \ac{goa} is limited to \{1280, 1024, 768, 512\} tokens respectively, while \ac{oora} is limited to \{1792, 1536, 1280, 1024\} tokens, reflecting the greater complexity of relation prediction.

\begin{table}[t]
  \centering
  \setlength{\tabcolsep}{3pt}
  \renewcommand{\arraystretch}{0.95}
  \caption{\textbf{Results for text-based \ac{lsa} (GT scene graphs, without SGG tool)}.
  Evaluation is based on Recall@K and meanRecall@K with $K=10,20,50$.
  We compare two settings:
  (1) \textbf{w/o \ac{goa}}: relation prediction conditioned on observed GT objects;
  (2) \textbf{w/ \ac{goa}}: future object prediction followed by per-object relation reasoning.
  \textbf{Bold} indicates the best result, and \underline{underline} indicates the second-best result.}
  \label{tab:pure_text_sga}
  \resizebox{\linewidth}{!}{%
  \begin{tabular}{l l c c c}
    \toprule
    \textbf{Setting} & \textbf{Method}
    & \textbf{R / mR@10} & \textbf{R / mR@20} & \textbf{R / mR@50} \\
    \midrule
    \multirow{6}{*}{\makecell{w/o \\\ac{goa}}}
    & GPT4o-mini                       & 47.2 / 32.6 & 49.2 / 38.2 & 49.2 / 38.3 \\
    & GPT4o                            & 61.2 / \textbf{43.7}
                                       & 64.6 / \underline{52.9}
                                       & 64.7 / \underline{53.2} \\
    & DeepSeek-V3                      & 55.2 / 38.9 & 57.9 / 46.4 & 58.0 / 46.6 \\
    & \cellcolor{blue!5}\textbf{\ac{ootsm}} \scriptsize{(DeepSeek-R1-1.5B)}
                                       & \cellcolor{blue!5}60.7 / 35.8
                                       & \cellcolor{blue!5}64.4 / 43.5
                                       & \cellcolor{blue!5}64.4 / 43.5 \\
    & \cellcolor{blue!5}\textbf{\ac{ootsm}} \scriptsize{(Qwen2.5-1.5B Instruct)}
                                       & \cellcolor{blue!5}62.0 / 36.2
                                       & \cellcolor{blue!5}65.1 / 43.7
                                       & \cellcolor{blue!5}65.1 / 43.7 \\
    & \cellcolor{blue!5}\textbf{\ac{ootsm}} \scriptsize{(Llama-3.2-3B Instruct)}
                                       & \cellcolor{blue!5}\textbf{68.7} / 42.8
                                       & \cellcolor{blue!5}\underline{72.4} / 51.8
                                       & \cellcolor{blue!5}\underline{72.4} / 51.8 \\
    \midrule
    \multirow{6}{*}{\makecell{w/ \\\ac{goa}}}
    & GPT4o-mini                       & 47.5 / 31.6 & 49.5 / 37.3 & 49.6 / 37.5 \\
    & GPT4o                            & 60.6 / \underline{43.3}
                                       & 63.9 / 52.5
                                       & 64.0 / 52.8 \\
    & DeepSeek-V3                      & 55.1 / 39.7 & 57.9 / 46.8 & 57.9 / 47.1 \\
    & \cellcolor{blue!5}\textbf{\ac{ootsm}} \scriptsize{(DeepSeek-R1-1.5B)}
                                       & \cellcolor{blue!5}61.0 / 35.3
                                       & \cellcolor{blue!5}65.6 / 44.2
                                       & \cellcolor{blue!5}65.6 / 44.2 \\
    & \cellcolor{blue!5}\textbf{\ac{ootsm}} \scriptsize{(Qwen2.5-1.5B Instruct)}
                                       & \cellcolor{blue!5}61.8 / 36.3
                                       & \cellcolor{blue!5}65.3 / 43.8
                                       & \cellcolor{blue!5}65.3 / 43.8 \\
    & \cellcolor{blue!5}\textbf{\ac{ootsm}} \scriptsize{(Llama-3.2-3B Instruct)}
                                       & \cellcolor{blue!5}\underline{68.4} / 41.5
                                       & \cellcolor{blue!5}\textbf{73.6} / \textbf{56.0}
                                       & \cellcolor{blue!5}\textbf{73.6} / \textbf{56.0} \\
    \bottomrule
  \end{tabular}%
  }
\end{table}
\begin{figure*}[t!]
    \centering
    \includegraphics[width=\linewidth]{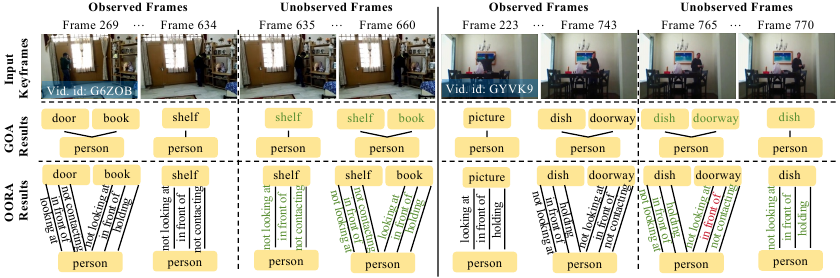}%
    \caption{\textbf{Qualitative results of \ac{ootsm}.}
    Two representative videos from Action Genome demonstrating newly appearing and disappearing objects in future predictions. Each example shows observed and unobserved frames with \ac{goa}-predicted future objects and \ac{oora}-anticipated relations. \textcolor{resultGreen}{Green} entries denote correct predictions; \textcolor{resultRed}{red} entries denote incorrect predictions.}
    \label{fig:qual_vis}
\end{figure*}

\subsection{Text‑based \ac{lsa} Results}
\label{exp:text_lsga_results}
\Cref{tab:pure_text_sga} reports Recall@K and meanRecall@K for text-based \ac{lsa} under two settings—w/o \ac{goa} (direct relation prediction, assuming all objects in the last observed frame persist) and w/ \ac{goa} (object forecasting before relation reasoning). In both cases, recall jumps markedly from R@10 to R@20, reflecting performance across short-, mid-, and long-term horizons.

Our proposed \ac{ootsm} (Llama-3.2-3B Instruct) achieves the highest R@20/50: 72.4\%/72.4\% without \ac{goa} and 73.6\%/73.6\% with \ac{goa}. GPT-4o is competitive, recording the highest mR@10 (43.7\%) without \ac{goa}. Adding predicted objects with \ac{goa} increases \ac{ootsm}'s mR@20 by 4.2\% over the single-stage setting, indicating that \ac{goa} refines the candidate object set and improves long tail prediction consistency.

\begin{table}[t]
  \centering
  \setlength{\tabcolsep}{3pt}
  \renewcommand{\arraystretch}{0.95}
  \caption{\textbf{Results for \ac{sga} on \ac{gags} under the with-constraint setting (using SGG tool).}
  Evaluation is conducted with different observed context fractions ($\mathcal{F}$).
  We report R / mR@10, R / mR@20, and R / mR@50.}
  \label{tab:sga_gags_constraint}
  \resizebox{\linewidth}{!}{%
  \begin{tabular}{c l c c c}
    \toprule
    \textbf{$\mathcal{F}$} & \textbf{Method}
    & \textbf{R / mR@10} & \textbf{R / mR@20} & \textbf{R / mR@50} \\
    \midrule

    \multirow{8}{*}{0.3}
    & STTran+                 & 30.8 / 7.1  & 32.8 / 7.8  & 32.8 / 7.8  \\
    & DSGDet+                 & 27.0 / 6.7  & 28.9 / 7.4  & 28.9 / 7.4  \\
    & STTran++                & 30.7 / 11.8 & 33.1 / 13.3 & 33.1 / 13.3 \\
    & DSGDet++                & 25.7 / 11.1 & 28.2 / 12.8 & 28.2 / 12.8 \\
    & SceneSayerODE           & 34.9 / 15.1 & 37.3 / 16.6 & 37.3 / 16.6 \\
    & SceneSayerSDE           & \textbf{39.7} / 18.4 & 42.2 / 20.5 & 42.3 / 20.5 \\
    & \cellcolor{blue!5}\textbf{\ac{ootsm} w/o \ac{goa} (Ours)}
                              & \cellcolor{blue!5}38.8 / \underline{19.5}
                              & \cellcolor{blue!5}\textbf{48.3} / \textbf{31.5}
                              & \cellcolor{blue!5}\textbf{49.3} / \textbf{33.7} \\
    & \cellcolor{blue!5}\textbf{\ac{ootsm} w/ \ac{goa} (Ours)}
                              & \cellcolor{blue!5}\underline{39.0} / \textbf{19.7}
                              & \cellcolor{blue!5}\underline{47.8} / \textbf{31.5}
                              & \cellcolor{blue!5}\underline{48.4} / \underline{32.3} \\
    \midrule

    \multirow{8}{*}{0.5}
    & STTran+                 & 35.0 / 8.0  & 37.1 / 8.7  & 37.1 / 8.8  \\
    & DSGDet+                 & 31.2 / 7.8  & 33.3 / 8.6  & 33.3 / 8.6  \\
    & STTran++                & 35.6 / 15.2 & 38.1 / 17.8 & 38.1 / 15.2 \\
    & DSGDet++                & 29.3 / 13.9 & 31.9 / 20.6 & 32.0 / 13.9 \\
    & SceneSayerODE           & 40.7 / 17.4 & 43.4 / 19.2 & 43.4 / 19.3 \\
    & SceneSayerSDE           & \textbf{45.0} / 20.7 & 47.7 / 23.0 & 47.7 / 23.1 \\
    & \cellcolor{blue!5}\textbf{\ac{ootsm} w/o \ac{goa} (Ours)}
                              & \cellcolor{blue!5}\underline{44.9} / \textbf{23.1}
                              & \cellcolor{blue!5}\underline{50.1} / \underline{29.9}
                              & \cellcolor{blue!5}\underline{50.1} / \underline{29.9} \\
    & \cellcolor{blue!5}\textbf{\ac{ootsm} w/ \ac{goa} (Ours)}
                              & \cellcolor{blue!5}43.4 / \underline{21.6}
                              & \cellcolor{blue!5}\textbf{52.3} / \textbf{33.8}
                              & \cellcolor{blue!5}\textbf{52.8} / \textbf{35.9} \\
    \midrule

    \multirow{8}{*}{0.7}
    & STTran+                 & 40.0 / 9.1  & 41.8 / 9.8  & 41.8 / 9.8  \\
    & DSGDet+                 & 35.5 / 9.6  & 37.3 / 9.6  & 37.3 / 9.6  \\
    & STTran++                & 41.3 / 18.2 & 43.6 / 18.2 & 43.6 / 18.2 \\
    & DSGDet++                & 33.9 / 15.9 & 36.3 / 15.9 & 36.3 / 15.9 \\
    & SceneSayerODE           & 49.1 / 21.0 & 51.6 / 22.9 & 51.6 / 22.9 \\
    & SceneSayerSDE           & 52.0 / 24.1 & 54.5 / 26.5 & 54.5 / 26.5 \\
    & \cellcolor{blue!5}\textbf{\ac{ootsm} w/o \ac{goa} (Ours)}
                              & \cellcolor{blue!5}\underline{52.6} / \underline{26.6}
                              & \cellcolor{blue!5}\underline{56.9} / \underline{32.9}
                              & \cellcolor{blue!5}\underline{56.9} / \underline{32.9} \\
    & \cellcolor{blue!5}\textbf{\ac{ootsm} w/ \ac{goa} (Ours)}
                              & \cellcolor{blue!5}\textbf{53.6} / \textbf{26.8}
                              & \cellcolor{blue!5}\textbf{58.9} / \textbf{34.8}
                              & \cellcolor{blue!5}\textbf{58.9} / \textbf{34.9} \\
    \midrule

    \multirow{8}{*}{0.9}
    & STTran+                 & 44.7 / 10.3 & 45.9 / 10.8 & 45.9 / 10.8 \\
    & DSGDet+                 & 38.8 / 10.2 & 40.0 / 10.7 & 40.0 / 10.7 \\
    & STTran++                & 46.0 / 19.6 & 47.7 / 21.4 & 47.7 / 21.4 \\
    & DSGDet++                & 38.1 / 16.3 & 39.8 / 17.7 & 39.8 / 17.7 \\
    & SceneSayerODE           & 58.1 / 25.0 & 59.8 / 26.4 & 59.8 / 26.4 \\
    & SceneSayerSDE           & 60.3 / 28.5 & 61.9 / 29.8 & 61.9 / 29.8 \\
    & \cellcolor{blue!5}\textbf{\ac{ootsm} w/o \ac{goa} (Ours)}
                              & \cellcolor{blue!5}\textbf{61.2} / \underline{31.3}
                              & \cellcolor{blue!5}\underline{69.8} / \underline{43.2}
                              & \cellcolor{blue!5}\underline{70.1} / \underline{43.8} \\
    & \cellcolor{blue!5}\textbf{\ac{ootsm} w/ \ac{goa} (Ours)}
                              & \cellcolor{blue!5}\underline{60.6} / \textbf{31.9}
                              & \cellcolor{blue!5}\textbf{73.2} / \textbf{48.1}
                              & \cellcolor{blue!5}\textbf{74.2} / \textbf{51.7} \\
    \bottomrule
  \end{tabular}%
  }
\end{table}
\subsection{Video-based \ac{sga} Results}
\label{exp:sga_results}
Results for video-based \ac{sga} under constrained conditions (\cref{tab:sga_gags_constraint}) demonstrate our \ac{ootsm} method's consistent superiority across all observation fractions ($\mathcal{F}=0.3,0.5,0.7,0.9$). When compared to strong baselines (\eg, SceneSayerSDE), our approach exhibits substantial performance advantages, particularly in medium-to-long-range predictions at R@20 and R@50. At $\mathcal{F}=0.9$, \ac{ootsm} achieves significant improvements of 11.3\% and 12.3\% in R@20 and R@50, respectively, relative to SceneSayerSDE. The pronounced enhancements in mR metrics further demonstrate our method's capacity for effective long-tail relation prediction through language prior integration. 

As illustrated in \cref{fig:qual_vis}, we present two representative cases: one video where new objects emerge in future frames, and another where existing objects gradually disappear. \ac{goa} predicts object set dynamics based on global context rather than mechanically replicating the last observed frame. This capability stems from \acp{llm}' ability to model temporal information in textual scene graph sequences, enabling them to learn semantically stable dynamic patterns and smooth local noise. Building on these predictions, \ac{oora} produces correct relationship predictions at most timesteps (marked in \textcolor{resultGreen}{green}), with errors (in \textcolor{resultRed}{red}) primarily occurring at fine-grained boundaries of attention/contact state transitions, demonstrating that the model effectively generates plausible temporal evolution following anticipated object trajectories.


\begin{table}[t]
  \centering
  \scriptsize
  \setlength{\tabcolsep}{4pt}
  \renewcommand{\arraystretch}{0.95}
  \caption{\textbf{Ablation on weighting and transition regularizers.}
  \textbf{Weight} refers to the cosine-weighted CE loss applied in \ac{goa}, and
  \textbf{Transition} denotes the transition loss introduced in \ac{oora}.}
  \label{tab:ablation_study}
  \resizebox{\linewidth}{!}{%
  \begin{tabular}{c c c c c c}
    \toprule
    \textbf{Index} & \textbf{Weighting} & \textbf{Transition}
    & \textbf{R / mR@10} & \textbf{R / mR@20} & \textbf{R / mR@50} \\
    \midrule
    (a) & —      & —      & 67.5 / 41.3 & 72.6 / 53.9 & 72.6 / 53.9 \\
    (b) & —      & \cmark & 68.1 / 41.4 & 73.3 / 54.9 & 73.3 / 54.9 \\
    (c) & \cmark & —      & 67.9 / \textbf{42.0}
                         & 73.3 / 55.6
                         & 73.3 / 55.6 \\
    \rowcolor{blue!5}
    (d) & \cmark & \cmark & \textbf{68.4} / 41.5
                         & \textbf{73.6} / \textbf{56.0}
                         & \textbf{73.6} / \textbf{56.0} \\
    \bottomrule
  \end{tabular}%
  }
\end{table}

\subsection{Impact of Cosine–Weighted and Transition Losses}\label{exp:abl_weight_trans}

We evaluate two temporal regularizers on the validation split.  
Cosine‑weighted CE loss is applied in \ac{goa}; each token loss is rescaled by the decay factor of~\cref{eq:cosine_weight} with $\beta=0.5$. 
Transition loss is the gated KL term introduced in \ac{oora} and weighted by $\lambda=0.03$.  
The BCE used to train the classifier head in \ac{oora} does not update the \ac{llm} backbone and is therefore omitted from this ablation.

The results are shown as~\cref{tab:ablation_study}. Adding the cosine weighting lifts overall recall from 72.6\% to 73.3\% and increases the long‑tail metric mR@20 by 1.7\% (c).  
The gains confirm that down‑weighting distant‑future tokens reduces noisy gradients and lets the model emphasize near‑term, high‑confidence patterns, markedly improving tail recall. Using the transition loss alone also reaches 73.3\% R@20, but improves mR@20 only to 54.9\% (b), suggesting it mainly enforces smoothness on frequent relations.  
Combining both regularizers (d) further raises overall recall to 73.6\% and yields the best balanced score of 56.0\% mR@20, showing that the two losses are complementary: cosine weighting broadens tail coverage, while the transition loss stabilises common relational patterns.  
The slight drop in mR@10 compared with row~c indicates that excessive smoothing can still dampen legitimate abrupt changes in very short windows, motivating future adaptive gating strategies.

\begin{table}[t]
  \centering
  \small
  \setlength{\tabcolsep}{3pt}
  \renewcommand{\arraystretch}{0.95}
  \caption{\textbf{One-shot and zero-shot performance on pure text-based \ac{lsa} (w/ \ac{goa}).}
  Each model first predicts future objects before performing relation reasoning.
  R@K and mR@K ($K=10,20,50$) are reported. \textbf{Bold} indicates the best result,
  and \underline{underline} indicates the second-best result.}
  \label{tab:one_shot_sga}
  \resizebox{\linewidth}{!}{%
  \begin{tabular}{l l c c c}
    \toprule
    \textbf{Mode} & \textbf{Method}
    & \textbf{R / mR@10} & \textbf{R / mR@20} & \textbf{R / mR@50} \\
    \midrule
    \multirow{3}{*}{One-shot}
    & GPT4o-mini
      & 60.4 / 41.4
      & 63.5 / 49.4
      & 63.5 / 49.6 \\
    & GPT4o
      & \underline{66.5} / \textbf{48.1}
      & \underline{70.3} / 48.7
      & \underline{70.4} / 48.9 \\
    & DeepSeek-V3
      & 60.2 / 41.8
      & 63.3 / 50.2
      & 63.3 / 50.5 \\
    \midrule
    \multirow{4}{*}{Zero-shot}
    & GPT4o-mini
      & 47.5 / 31.6
      & 49.5 / 37.3
      & 49.6 / 37.5 \\
    & GPT4o
      & 60.6 / \underline{43.3}
      & 63.9 / \underline{52.5}
      & 64.0 / \underline{52.8} \\
    & DeepSeek-V3
      & 55.1 / 39.7
      & 57.9 / 46.8
      & 57.9 / 47.1 \\
    & \cellcolor{blue!5}\ac{ootsm} \scriptsize{(Llama-3.2-3B Instruct)}
      & \cellcolor{blue!5}\textbf{68.4} / 41.5
      & \cellcolor{blue!5}\textbf{73.6} / \cellcolor{blue!5}\textbf{56.0}
      & \cellcolor{blue!5}\textbf{73.6} / \cellcolor{blue!5}\textbf{56.0} \\
    \bottomrule
  \end{tabular}%
  }
\end{table}

\subsection{Effectiveness of Temporal Reasoning}
\label{exp:temporal_reasoning}

\begin{table}[t]
  \centering
  \small
  \setlength{\tabcolsep}{4pt}
  \renewcommand{\arraystretch}{0.95}
  \caption{\textbf{Validation of temporal reasoning effectiveness across observation ratios.}
  $^{\dagger}$ denotes the naive strategy (predicting all future frames using the last observed frame’s scene graph).
  We report R / mR@10, R / mR@20, and R / mR@50.}
  \label{tab:diagnostic_sga_evaluation}
  \resizebox{\linewidth}{!}{%
  \begin{tabular}{c l c c c}
    \toprule
    \textbf{$\mathcal{F}$} & \textbf{Method}
    & \textbf{R / mR@10} & \textbf{R / mR@20} & \textbf{R / mR@50} \\
    \midrule

    \multirow{3}{*}{0.3}
    & SceneSayerSDE$^{\dagger}$ & 38.7 / 16.6 & 42.0 / 19.9 & 42.1 / 19.9 \\
    & SceneSayerSDE             & 39.7 / 18.4 & 42.2 / 20.5 & 42.3 / 20.5 \\
    & \cellcolor{blue!5}\ac{ootsm}                     & \cellcolor{blue!5}39.0 / 19.7
                                & \cellcolor{blue!5}47.8 / 31.5
                                & \cellcolor{blue!5}48.4 / 32.3 \\
    \midrule

    \multirow{3}{*}{0.5}
    & SceneSayerSDE$^{\dagger}$ & 43.4 / 18.5 & 47.1 / 21.9 & 47.1 / 22.1 \\
    & SceneSayerSDE             & 45.0 / 20.7 & 47.7 / 23.0 & 47.7 / 23.1 \\
    & \cellcolor{blue!5}\ac{ootsm}                     & \cellcolor{blue!5}43.4 / 21.6
                                & \cellcolor{blue!5}52.3 / 33.8
                                & \cellcolor{blue!5}52.8 / 35.9 \\
    \midrule

    \multirow{3}{*}{0.7}
    & SceneSayerSDE$^{\dagger}$ & 50.7 / 21.9 & 53.9 / 25.2 & 53.9 / 25.2 \\
    & SceneSayerSDE             & 52.0 / 24.1 & 54.5 / 26.5 & 54.5 / 26.5 \\
    & \cellcolor{blue!5}\ac{ootsm}                     & \cellcolor{blue!5}53.6 / 26.8
                                & \cellcolor{blue!5}58.9 / 34.8
                                & \cellcolor{blue!5}58.9 / 34.9 \\
    \midrule

    \multirow{3}{*}{0.9}
    & SceneSayerSDE$^{\dagger}$ & 59.3 / 26.5 & 61.5 / 29.2 & 61.5 / 29.2 \\
    & SceneSayerSDE             & 60.3 / 28.5 & 61.9 / 29.8 & 61.9 / 29.8 \\
    & \cellcolor{blue!5}\ac{ootsm}                     & \cellcolor{blue!5}60.6 / 31.9
                                & \cellcolor{blue!5}73.2 / 48.1
                                & \cellcolor{blue!5}74.2 / 51.7 \\
    \midrule

    \multirow{2}{*}{\makecell{Avg $\Delta$ \\ \scriptsize{vs SceneSayerSDE$^{\dagger}$}}}
    & SceneSayerSDE             & \textbf{+1.3} / +2.0 & +0.5 / +0.9 & +0.4 / +0.9 \\
    & \cellcolor{blue!5}\ac{ootsm}                     & \cellcolor{blue!5}+1.2 / \textbf{+4.1}
                                & \cellcolor{blue!5}\textbf{+7.0} / \textbf{+13.0}
                                & \cellcolor{blue!5}\textbf{+7.4} / \textbf{+14.6} \\
    \bottomrule
  \end{tabular}%
  }
\end{table}

To verify that \ac{ootsm} learns genuine temporal dynamics rather than relying on final-frame replication, we evaluate across observation ratios $\mathcal{F} \in \{0.3, 0.5, 0.7, 0.9\}$ (Table~\ref{tab:diagnostic_sga_evaluation}). \ac{ootsm} demonstrates substantially larger improvements in mR metrics than R metrics, indicating better performance on long-tail categories and newly appearing objects. At $\mathcal{F}=0.9$, mR improvements (18.3\% and 21.9\% for mR@20 and mR@50) exceed R improvements (11.3\% and 12.3\%), with average gains of +4.1\% (mR@10), +13.0\% (mR@20), and +14.6\% (mR@50) over the naive baseline. This suggests our approach leverages linguistic priors to predict infrequent relationships that visual methods typically miss.

Performance gains increase substantially at larger K values, with R@20 and R@50 improvements (+7.0\% and +7.4\%) exceeding short-term gains (+1.2\% at R@10). This pattern indicates that linguistic reasoning is more effective for longer horizons, where structured scene graphs provide cleaner signals than raw visual features, enabling \acp{llm} to better capture temporal object dynamics.

\subsection{One-Shot Prompting Strategy}
\label{exp:one_shot_results}

We evaluate API-based \acp{llm} on pure text-based \ac{lsa} under zero-shot and one-shot prompting conditions (\cref{tab:one_shot_sga}). Incorporating a single demonstration example generally enhances API model performance: GPT4o-mini improves from 47.5\% to 60.4\% (R@10) and from 31.6\% to 41.4\% (mR@10), while DeepSeek-V3 shows consistent gains across metrics. GPT4o achieves the best one-shot R@10 performance (66.5\%), though its class-balanced metrics (mR@20, mR@50) slightly degrade compared to zero-shot, suggesting potential trade-offs between overall and tail-class accuracy.

In contrast, our fine-tuned \ac{ootsm} model achieves superior zero-shot performance, surpassing one-shot results of competing models across most metrics. Notably, \ac{ootsm} reaches 73.6\% R@20 and 56.0\% mR@20 in zero-shot settings, demonstrating the effectiveness of task-specific fine-tuning for \ac{lsa} without requiring in-context examples.

\subsection{Effect of the Observation Window}
\label{sec:obs_len}
\begin{figure}[t!]
  \centering
  \captionsetup[subfloat]{font=footnotesize}
  \subfloat[Observation window.\label{fig:obs_len_curve}]{
    \includegraphics[width=1\linewidth]{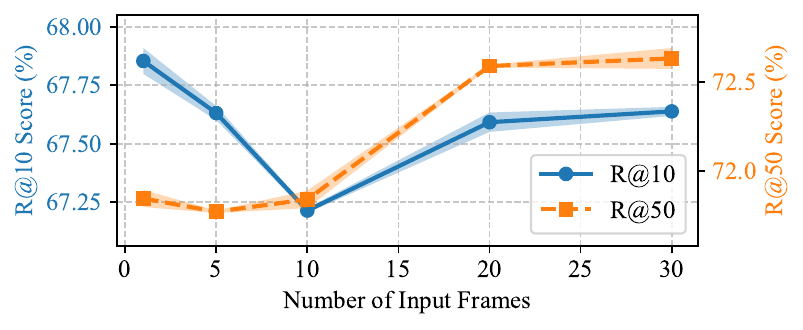}
  }\hfill
  \subfloat[Cosine weighting ($\beta$).\label{fig:beta_curve}]{
    \includegraphics[width=1\linewidth]{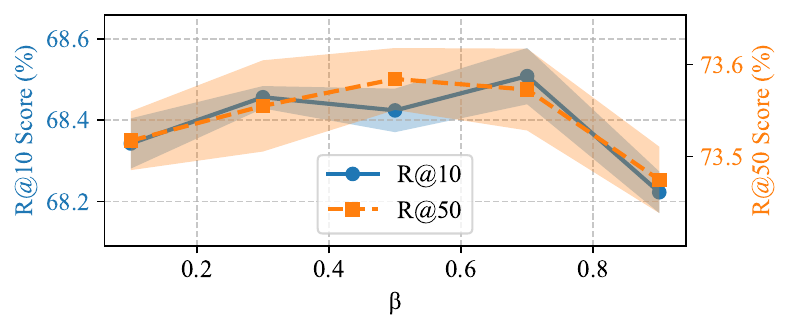}
  }
  \caption{\textbf{Observation window length and weighting hyper-parameter.}
    Left: effect of varying the number of observed frames on recall performance.
    Right: study of the cosine-weighting coefficient $\beta$. Shaded region denotes
    one standard deviation over three independent runs per value.}
  \label{fig:obs_len_vs_perf}
\end{figure}

\Cref{fig:obs_len_curve} plots short‐term (R@10) and long‐term (R@50) prediction accuracy as the number of input observation frames increases from 1 to 30. When only a few frames are observed, the model achieves its highest R@10 but exhibits relatively poor R@50. As more frames are added, R@50 recovers significantly and then plateaus, while R@10, after a small initial dip, gradually returns to its original level. In summary, adding history frames yields clear gains in long‐range prediction but diminishing returns for short‐range accuracy. Nevertheless, to maximize available temporal context, we feed as many frames as possible—subject to a 2k token limit—to strike a balance between predictive performance and inference latency.

\subsection{Effect of the Weighting Hyper-Parameter}
\label{exp:weighting}
As shown in~\cref{fig:beta_curve}, we examine \ac{goa}'s training sensitivity to the cosine-weighting coefficient $\beta$, which modulates future token loss down-weighting. The performance curve exhibits a shallow U-shape across prediction horizons. Very small $\beta$ values concentrate loss on near-term frames, causing overfitting to short-range patterns and degrading overall performance. Moderate $\beta$ values balance high-confidence short-range supervision with sufficient long-range exposure to regularize temporal dynamics, yielding monotonically improving recall. Beyond optimal levels, near-uniform weighting reintroduces noisy gradients from uncertain far-future tokens, reverting metrics toward their low-$\beta$ baseline. Our findings indicate that moderate decay ($0.3<\beta<0.7$) optimally balances near-future reliability with comprehensive horizon coverage.
\subsection{Robustness Analysis}
\label{exp:robutness_analysis}

\begin{table}[t]
  \centering
  \small
  \setlength{\tabcolsep}{4pt}
  \renewcommand{\arraystretch}{0.95}
  \caption{\textbf{Robustness to injected SGG noise.}
  Each cell reports short-term / long-term performance (R@10 / R@50) under different noise types, frame ranges, and noise ratios.
  5\%--30\% indicates the proportion of frames replaced by noise.}
  \label{tab:noise-robust}
  \resizebox{\linewidth}{!}{%
  \begin{tabular}{l c c c c c}
  \toprule
  \multirow{2}{*}{\makecell{\textbf{Noisy}\\\textbf{Type}}} &
  \multirow{2}{*}{\makecell{\textbf{Frame}\\\textbf{Range}}} &
  \textbf{5\%} &
  \textbf{10\%} &
  \textbf{15\%} &
  \textbf{30\%} \\
  \cmidrule(lr){3-6}
  & & \scriptsize \textbf{R@10 / R@50}
    & \scriptsize \textbf{R@10 / R@50}
    & \scriptsize \textbf{R@10 / R@50}
    & \scriptsize \textbf{R@10 / R@50} \\
  \midrule
    drop   & 0--30\%  & 68.2 / 73.4 & 68.3 / 73.4 & 68.2 / 73.5 & 68.5 / 73.6 \\
    drop   & 30--60\% & 68.4 / 73.4 & 68.4 / 73.5 & 68.4 / 73.4 & 68.3 / 73.5 \\
    drop   & 60--90\% & 68.6 / 73.6 & 68.5 / 73.7 & 68.7 / 73.8 & 68.8 / 73.9 \\
    \midrule
    modify & 0--30\%  & 67.2 / 73.1 & 67.1 / 73.2 & 66.6 / 73.0 & 66.4 / 73.3 \\
    modify & 30--60\% & 66.8 / 73.3 & 65.5 / 73.2 & 64.8 / 72.6 & 63.4 / 72.7 \\
    modify & 60--90\% & 66.2 / 73.4 & 64.7 / 73.0 & 63.5 / 72.9 & 62.0 / 71.6 \\
    \midrule
    Avg $\Delta$ & &
      \cellcolor{blue!5}-1.22 / -0.32 &
      \cellcolor{blue!5}-1.92 / -0.36 &
      \cellcolor{blue!5}-2.49 / -0.54 &
      \cellcolor{blue!5}-3.16 / -0.68 \\
    \bottomrule
  \end{tabular}%
  }
\end{table}

We evaluate robustness to \ac{sgg} noise and detector quality variations. \Cref{tab:noise-robust} shows that \ac{ootsm} demonstrates high tolerance to missing detections (drop noise), maintaining stable performance across all noise rates with R@50 degradation below 0.7\%  even at 30\% noise. For incorrect detections (modify noise), short-term performance (R@10) degrades proportionally with noise rate, while long-term predictions (R@50) remain remarkably robust, degrading by only 0.68\% on average at 25\% frame error rates. Temporal analysis reveals that noise in later frames has greater impact on immediate predictions, quantifying the model's reliance on recent observations.

\begin{table}[t]
  \centering
  \small
  \setlength{\tabcolsep}{3pt}
  \renewcommand{\arraystretch}{1.0}
  \caption{\textbf{Performance improvements with different SGG backbones.}
  \ac{ootsm} consistently enhances both weaker and stronger detectors across all metrics, with particularly notable gains in class-balanced recall (mR).}
  \label{tab:sgg-tools}
  \resizebox{\linewidth}{!}{%
  \begin{tabular}{l c c c c}
    \toprule
    \textbf{Method} & \textbf{R / mR@10} & \textbf{R / mR@20} & \textbf{R / mR@50} & \textbf{$\Delta$ mR@50} \\
    \midrule
    DSGDet+                    & 38.8 / 10.2 & 40.0 / 10.7 & 40.0 / 10.7 & -- \\
    \cellcolor{blue!5}+ \ac{ootsm}    
                              & \cellcolor{blue!5}60.5 / 25.7
                              & \cellcolor{blue!5}64.6 / 29.6
                              & \cellcolor{blue!5}64.6 / 29.6
                              & \cellcolor{blue!5}\textbf{+18.9} \\
    \midrule
    DSGDet++                   & 38.1 / 16.3 & 39.8 / 17.7 & 39.8 / 17.7 & -- \\
    \cellcolor{blue!5}+ \ac{ootsm}    
                              & \cellcolor{blue!5}59.5 / 29.2
                              & \cellcolor{blue!5}65.7 / 34.7
                              & \cellcolor{blue!5}65.7 / 34.8
                              & \cellcolor{blue!5}\textbf{+17.1} \\
    \midrule
    STTran+                    & 44.7 / 10.3 & 45.9 / 10.8 & 45.9 / 10.8 & -- \\
    \cellcolor{blue!5}+ \ac{ootsm}    
                              & \cellcolor{blue!5}60.0 / 23.7
                              & \cellcolor{blue!5}63.9 / 27.6
                              & \cellcolor{blue!5}64.0 / 27.7
                              & \cellcolor{blue!5}\textbf{+16.9} \\
    \midrule
    STTran++                   & 46.0 / 19.6 & 47.7 / 21.4 & 47.7 / 21.4 & -- \\
    \cellcolor{blue!5}+ \ac{ootsm}    
                              & \cellcolor{blue!5}60.6 / 31.9
                              & \cellcolor{blue!5}73.2 / 48.1
                              & \cellcolor{blue!5}74.2 / 51.7
                              & \cellcolor{blue!5}\textbf{+30.3} \\
    \bottomrule
  \end{tabular}%
  }
\end{table}

We further evaluate compatibility with different \ac{sgg} tools (seen in \cref{tab:sgg-tools}). \ac{ootsm} consistently improves performance across detectors of varying quality. Notably, adding \ac{ootsm} to the stronger STTran++ backbone achieves the most substantial gain of +30.3\% mR@50, while even weaker detectors like DSGDet+ show +18.9\% mR@50 improvement. These results demonstrate that \ac{ootsm} maintains effectiveness across diverse \ac{sgg} quality levels, validating its practical applicability in real-world scenarios.

\section{Conclusion}
\label{sec:conclusion}
This work introduces \ac{ootsm}, a novel two-stage, language-driven framework that fundamentally reframes scene-graph anticipation as textual inference. Our fine-tuned \ac{llm} with cosine-weighted and transition loss demonstrates state-of-the-art performance by consistently outperforming visual baselines on both our pioneering text-only \ac{lsa} benchmark and standard \ac{sga} evaluation, particularly at extended prediction horizons. These advantages stem from the principled separation of object and relation reasoning within a linguistic interface that naturally produces unordered multi-label triples, accommodates object dynamics, and effectively leverages linguistic associations when visual evidence is ambiguous or incomplete. Future directions include developing end-to-end video integration that eliminates modality transitions, implementing open-vocabulary object representation to address long-tail recognition failures and enhance generalization across varied scenarios, and exploring adaptive temporal constraints to balance between stability and legitimate state changes.
\section*{Acknowledgement}
\label{sec:acknowledgement}
The authors would like to thank Yuyang Li, Fengming Zhu, and Yangfan Wu for their valuable discussions and constructive feedback that helped improve this work.

\setstretch{0.975}
\bibliography{reference_header,reference}
\bibliographystyle{ieeetr}
\setstretch{1}

\vskip -3\baselineskip plus -1fil
\begin{IEEEbiography}[{\includegraphics[width=1in,height=1.25in,clip,keepaspectratio]{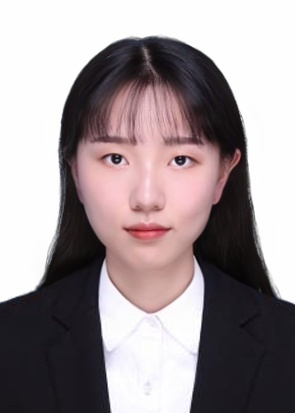}}]{Xiaomeng Zhu} is a second-year Ph.D. student in the Department of Computer Science and Engineering at the Hong Kong University of Science and Technology. She received her B.Eng. degree in automation from the University of Electronic Science and Technology of China and her Master's degree in pattern recognition and intelligent systems from the Institute of Automation, Chinese Academy of Sciences. Her research focuses on advancing robotic understanding of human intent and environmental context to enable more effective human-robot collaboration.
\end{IEEEbiography}

\begin{IEEEbiography}[{\includegraphics[width=1in,height=1.25in,clip,keepaspectratio]{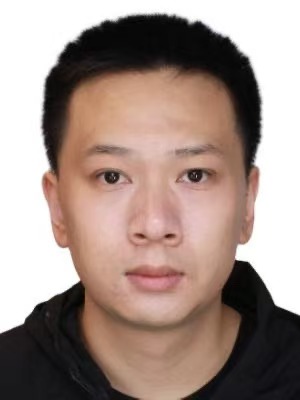}}]{Changwei Wang}
received the B.S. degree in software engineering from Tiangong University, China, in July 2019.  He received the Ph.D degree in the State Key Laboratory of Multimodal Artificial Intelligence Systems, Institute of Automation, Chinese Academy of Sciences and School of Artificial Intelligence, University of Chinese Academy of Sciences, China, in July 2024. From 2024, he works at Shandong Computer Science Center (National Supercomputer Center in Jinan). 
His research interests include multimodal learning, robotic vision, embodied intelligence and edge intelligence.
\end{IEEEbiography}

\begin{IEEEbiography}[{\includegraphics[width=1in,height=1.25in,clip,keepaspectratio]{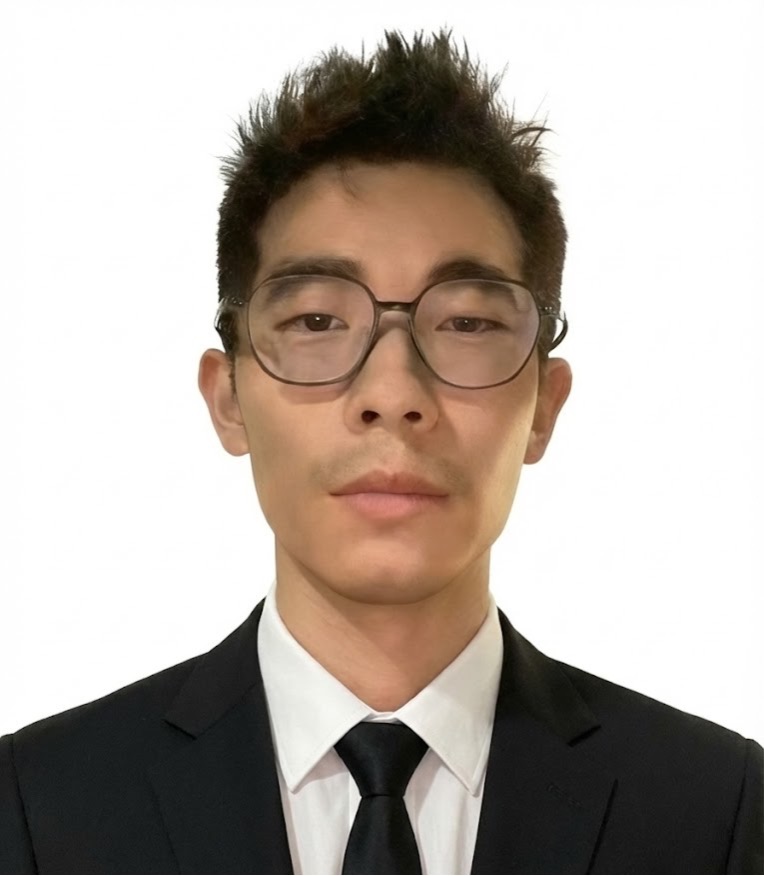}}]{Haozhe Wang} is a second-year PhD student with the Hong Kong University of Science and Technology (HKUST), where he is a recipient of the prestigious Hong Kong PhD Fellowship Scheme (HKPFS). His research interests focus on reasoning and Reinforcement Learning (RL) for Large Language Models (LLMs) and Vision-Language Models (VLMs). He has published over ten papers in top-tier conferences, including NeurIPS, AAAI, ACL, etc. He also serves as a Program Committee member for premier venues such as NeurIPS, ICLR, CVPR, and TMLR.
\end{IEEEbiography}
\begin{IEEEbiography}[{\includegraphics[width=1in,height=1.25in,clip,keepaspectratio]{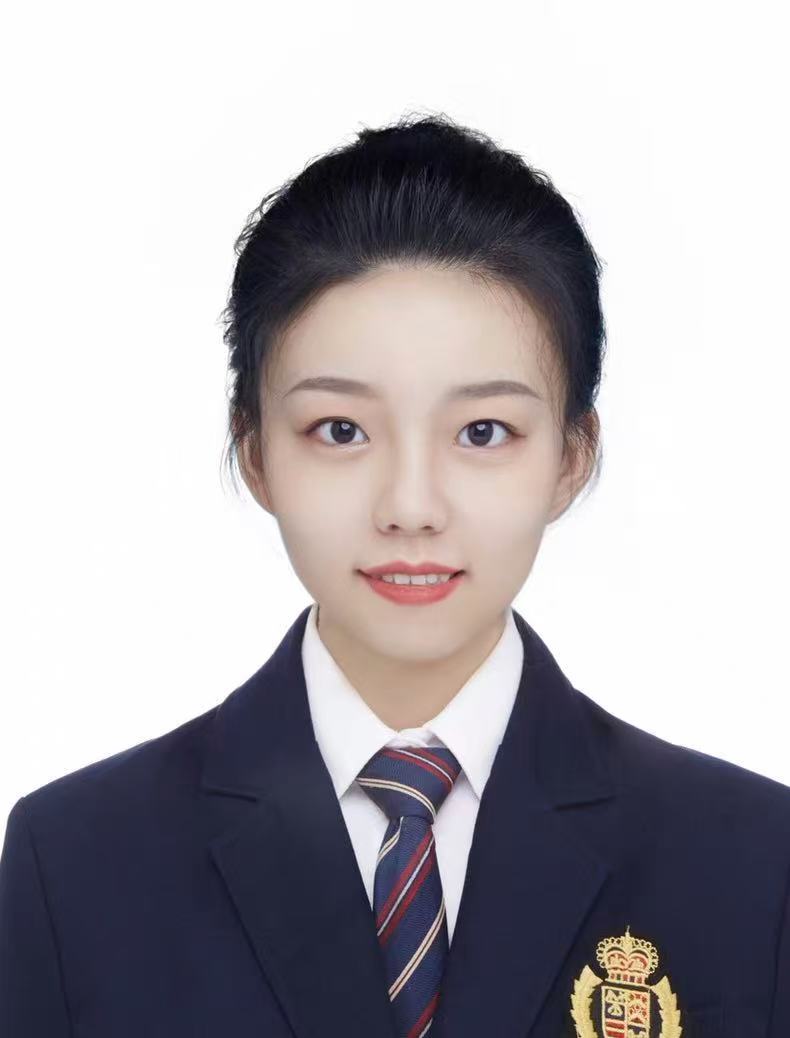}}]{Xinyu Liu} is a second-year Ph.D. student in the Academy of Interdisciplinary Studies at the Hong Kong University of Science and Technology. She received a Bachelor's degree in Communication Engineering from Xidian University and a Master's degree in Computer Science and Technology from the School of Artificial Intelligence at Xidian University. Her research focuses on the understanding and generation of dynamic world context.
\end{IEEEbiography}

\begin{IEEEbiography}[{\includegraphics[width=1in,height=1.25in,clip,keepaspectratio]{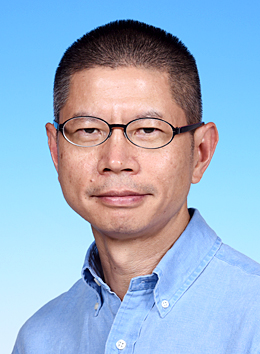}}]{Fangzhen Lin} is currently a Professor in the Department of Computer Science and Engineering of the Hong Kong University of Science and Technology. He received his BS degree at Fuzhou University, MS degree at Beijing University and PhD at Stanford University. His research interests is in artificial intelligence, and in particular, principles of knowledge representation, reasoning, and learning and their applications in programming languages, robotics, multi--agent systems, game theory and social choice theory, language understanding etc. He is an AAAI Fellow and received a Croucher Senior Research Fellowship.
\end{IEEEbiography}
\vfill

\end{document}